%% file: main.tex
\documentclass[letterpaper, 10 pt, conference]{ieeeconf}
\IEEEoverridecommandlockouts
\usepackage{times}

\usepackage{multicol}
\usepackage[bookmarks=true]{hyperref}
\usepackage{color}
\usepackage[utf8]{inputenc}

\usepackage{amsthm,amsmath,amssymb}
\usepackage{graphicx}
\usepackage{import}
\usepackage{dsfont}
\usepackage{todonotes}
\usepackage{amsfonts}
\usepackage{algorithm}
\usepackage{mathtools}
\usepackage{acronym}
\usepackage{algpseudocode} 
\usepackage{tabularx}
\usepackage[most]{tcolorbox}
\usepackage{diagbox}
\usepackage[utf8]{inputenc}
\usepackage[export]{adjustbox}

\usepackage{svg}
\usepackage{cite}

\let\labelindent\relax
\usepackage{enumitem}
\usepackage{makecell}
\usepackage{physics}

\title{\LARGE \bf
A Hybrid Model and Learning-Based Force Estimation Framework for Surgical Robots}
\author{Hao Yang, Haoying Zhou, Gregory S. Fischer and Jie Ying Wu%
\thanks{This work is supported in part by an Intuitive Surgical Technology Research Grant and NSF AccelNet award OISE-1927275. Hao Yang and Jie Ying Wu are with the Department of Computer Science, Vanderbilt University, TN 37212, USA. Haoying Zhou and Gregory S. Fischer are with the Department of Robotics Engineering, Worcester Polytechnic Institute, MA 01609, USA.
All correspondence should be addressed to Hao Yang {\tt\small hao.yang@vanderbilt.edu}}
}

\begin{document}




\maketitle
\thispagestyle{empty}
\pagestyle{empty}

\begin{abstract}

Haptic feedback to the surgeon during robotic surgery would enable safer and more immersive surgeries but estimating tissue interaction forces at the tips of robotically controlled surgical instruments has proven challenging. Few existing surgical robots can measure interaction forces directly and the additional sensor may limit the life of instruments. We present a hybrid model and learning-based framework for force estimation for the Patient Side Manipulators (PSM) of a da Vinci Research Kit (dVRK). The model-based component identifies the dynamic parameters of the robot and estimates free-space joint torque, while the learning-based component compensates for environmental factors, such as the additional torque caused by trocar interaction between the PSM instrument and the patient's body wall. We evaluate our method in an abdominal phantom and achieve an error in force estimation of under 10\% normalized root-mean-squared error. We show that by using a model-based method to perform dynamics identification, we reduce reliance on the training data covering the entire workspace. Although originally developed for the dVRK, the proposed method is a generalizable framework for other compliant surgical robots. The code is available at \url{https://github.com/vu-maple-lab/dvrk_force_estimation}.
    
\end{abstract}





\section{Introduction}
\label{sec:introduction}
\input{section/1.Introduction}

\section{Related Works}
\label{sec:related}
\input{section/2.Related_Works}

\section{Methods}
\label{sec:force_est}
\input{section/3.Force_est}

\section{Experimental Setup}
\label{sec:experimental_setup}
\input{section/4.Experimental_setup}

\section{Results and Discussion}
\label{sec:experiment}
\input{section/5.Experiment}

\section{Conclusion}
\label{sec:conclusion}
\input{section/6.Conclusion}


\bibliographystyle{IEEEtran}
\bibliography{references}

\end{document}

%% file: section/1.Introduction.tex

Robot-assisted minimally invasive surgery systems have revolutionized the laparoscopic surgery procedures. They provide an ergonomic setup for the surgeon while improving patient recovery time~\cite{Williamson2022RoboticST}. However, the lack of direct haptic feedback on all but a few select instruments is a significant barrier for surgeons in building surgical intuition. To enable haptic feedback on surgical robots more widely, various approaches have been suggested~\cite{Okamura2004MethodsFH, Okamura2007HapticsFR, Fischer2006IschemiaAF} that include integrating force sensors~\cite{Chua2022ModularForce, Fontanelli2020AnEF} and sensorless force estimation~\cite{Yilmaz2020NeuralNB, Wu2021RobotFE}. 

Sensor-based methods have the advantage of measuring forces independently of robot dynamics. Chua et al. propose a modular 3 degrees-of-freedom force sensor to measure grip force~\cite{Chua2022ModularForce}. Fontanelli et al. propose an optic-based force sensor, which can be integrated with a trocar and sense the interactive force between the instrument shaft and patient body~\cite{Fontanelli2020AnEF}. These sensor-based methods, however, require extra mechatronics, as well as extra manufacturing and sensor fusion efforts.

Another direction for providing haptic feedback is sensorless force estimation through calibrating robot models. Liu et al. develop a sensorless model-based scheme to identify system dynamics and estimate contact force for an industrial serial robot~\cite{Liu2021SensorlessFE}. However, accurate dynamic models for cable-driven surgical robots are complex as they contain parallelogram mechanisms, counterweight, springs, and cable friction. Additionally, the components of robots are subject to deformation and wear and tear along their life cycle. Moreover, during surgeries, the surgical robots move about the remote center of motion (RCM)~\cite{Eldridge1996ARC} through a trocar inserted into the patient's abdomen to guide the instrument. The instruments are then inserted into the abdominal cavity through the trocar. The interaction force between the patient body wall, the trocar, and the instrument is hard to model, which complicates in vivo force estimation~\cite{Deeken2017MechanicalPO}. Therefore, it is essential to develop force estimation methods that can provide an accurate dynamic model that can account for these interactions.

Both sensor-based and model-based approaches face challenges in widespread adoption. Force sensors are expensive and difficult to sterilize while model-based force estimation methods are limited. On the other hand, recent learning-based force estimation methods are highly dependent on training data and generalize poorly~\cite{Yilmaz2022TransferOL,Wu2021RobotFE,Yilmaz2020NeuralNB}. In this work, we propose a hybrid approach to force estimation that combines the generalizability of model-based method with the adaptability of the learning-based method. Specifically, we propose a hybrid model and learning-based joint torque estimation method and the Jacobian to convert joint torques to Cartesian forces. We first perform an optimization-based dynamic parameter identification and joint torque estimation of a dVRK~\cite{kazanzides-chen-etal-icra-2014}. Then, we add a neural network second step to account for extra joint torques caused by setup-specific trocar interaction that can be trained with minimal data.






%% file: section/2.Related_works.tex
\begin{figure*}[htbp]
    \centering
    \vspace{3mm}
    \includegraphics[width=0.85\linewidth]{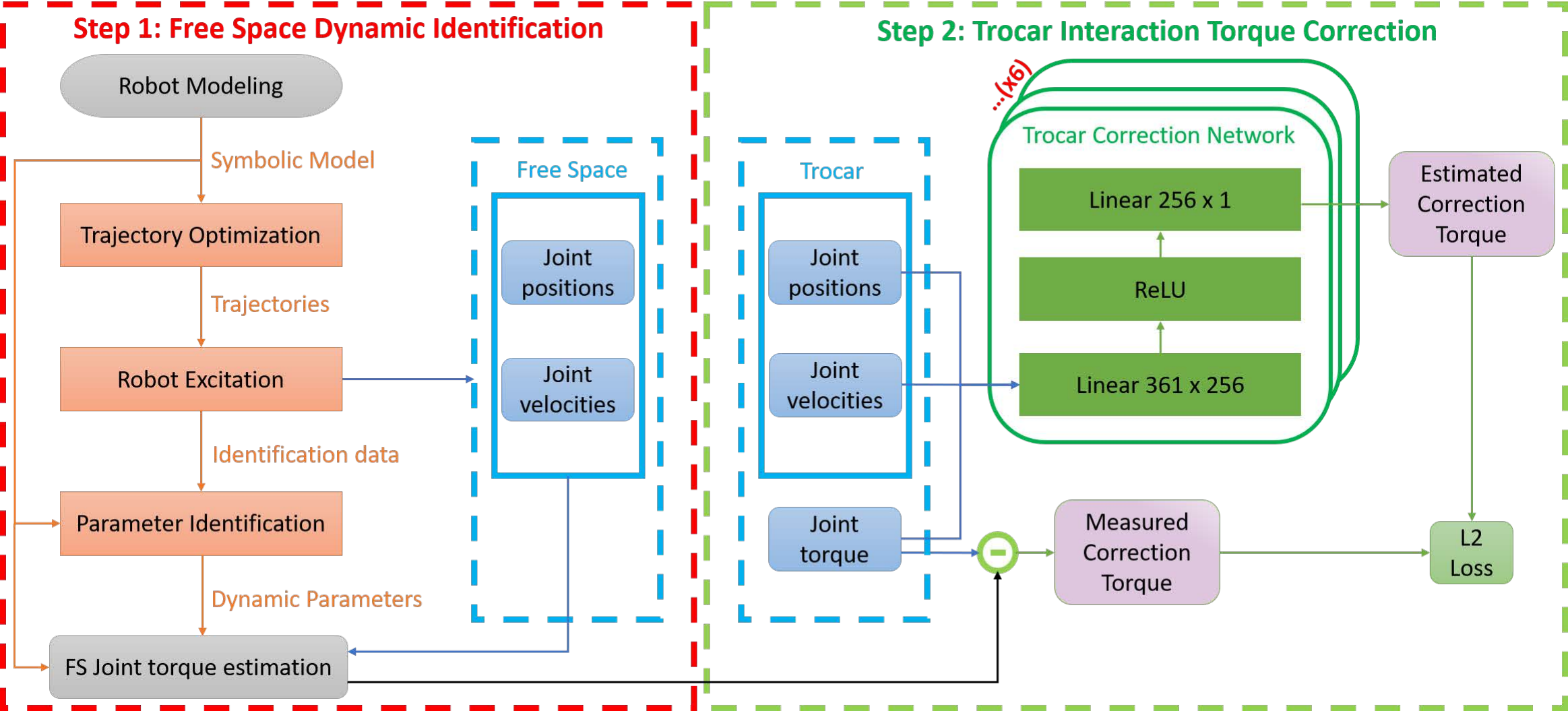}
    \caption{Block diagrams of the proposed two-step hybrid joint torque estimation method. The first step, shown on the left, is the model-based dynamic parameter identification and free space torque estimation process. The second step, shown on the right, is the learning-based torque correction scheme about in-trocar interaction.}
    \label{fig:tq_pred}
\end{figure*}

\subsection{Model-based Torque Estimation}
One class of dynamic identification and torque estimation methods is model-based. Sang et al. and Piqué et al. propose explicit dynamic models for the da Vinci Research Kit (dVRK) Patient Side Manipulator (PSM), with the free space dynamic parameters identified through ordinary least square optimization~\cite{Sang2017ExternalFE, pique2019dynamic}. Souza et. al state that in order to achieve a physically feasible robot, the mass inertia matrix in the inverse dynamic equation needs to be positive semidefinite and the mass of each link should always be positive~\cite{Sousa2014PhysicalFO}. Fontanelli et al. use these constraints to form a linear matrix inequality-based method for dynamic parameter and joint torque identification for dVRK PSM~\cite{Fontanelli2017ModellingAI}. On top of this, Wang et al. develop a convex optimization-based dynamic model identification package that estimates the free space torque of the dVRK~\cite{wang2019convex}. These methods construct explicit dynamic models and identify the dynamic parameters from free space torque. All these methods work well in free space dVRK torque estimation at the tooltip. However, none of them can address the challenge created by multiple points of interaction between the instrument shaft, trocar, and patient body.

\subsection{Learning-based Torque Estimation}
Alternatively, Yilmaz et al. propose a neural network (NN) based method to estimate the external force~\cite{Yilmaz2020NeuralNB}.  Wu et al. extend the work by adding a correction network for trocar interactions on an abdominal phantom~\cite{Wu2021RobotFE}. Further work leverages transfer learning to adapt pseudo-clinical settings, such that force estimation can be applied across different robots, different instruments, and different pivot points on an abdominal phantom~\cite{Yilmaz2022TransferOL}. However, our experiments show that these learning-based methods are highly reliant on training data quality to ensure estimation accuracy. Mismatch in the training and test workspaces can result in high errors. 

%% file: section/3.Force_est.tex

The hybrid torque estimation method proposed in this paper has two steps, which are shown in Fig.~\ref{fig:tq_pred}. Step one is model-based dynamic identification and free space torque estimation. We move the robot along an optimized excitation trajectory without any contact and optimize the robot's dynamic parameters. Step two is learning-based trocar interaction torque compensation. We train a NN to learn patient and setup specific effects. The NN is designed to be trained with a subset of the workspace that is relevant to the procedure after the instruments have engaged and passed through the ports. Then, we validate the effectiveness of this method by doing Cartesian force estimation, where the robot instrument tip is in contact with external objects while the shaft is inserted through the trocar and incision port.


\subsection{Model-based Free Space Dynamic Identification and Torque Estimation}

To address free space joint torque estimation, we first aim to use the dynamic parameter identification method proposed in~\cite{wang2019convex, pique2019dynamic, ferro2022coppeliasim} to obtain the dynamic model of the patient-side robot. The dynamic identification can be achieved through the following five steps:
\begin{enumerate}[label=(\alph*)]
    \item Build the symbolic kinematic model of the PSM based on the modified DH parameters~\cite{khalil1986new} of the PSM
    \item Construct the symbolic dynamic model of the PSM using Euler-Lagrangian approach~\cite{nakamura1989dynamics, spong2020robot}
    \item Calculate the optimized excitation trajectory based on the given joint constraints and the dynamic model
    \item Run the excitation trajectory on the physical PSM and collect kinematic and dynamic data through dVRK
    \item Solve for the model parameters using the least square and optimization methods
\end{enumerate}

\subsubsection{Kinematic Modeling of the PSM}

To build the relationship between the robot joint motion and the torque of each motor, we define four different types of joint coordinates:
\begin{itemize}
    \item $\mathbf{q^d}$: the joint coordinates obtained from the dVRK~\cite{kazanzides-chen-etal-icra-2014} package with coupled wrist joints
    \item $\mathbf{q}$: the joint coordinates used in the kinematic modeling with uncoupled wrist joints
    \item $\mathbf{q^m}$: the equivalent motor coordinates at joints
    \item $\mathbf{q^c}$: the complete joint coordinates, defined as $\mathbf{q^c} = [\mathbf{q}^T \, (\mathbf{q^m})^T]^T$
\end{itemize}

\begin{figure}[]
    \centering
    \vspace{2mm}
    \includegraphics[width=0.9\linewidth]{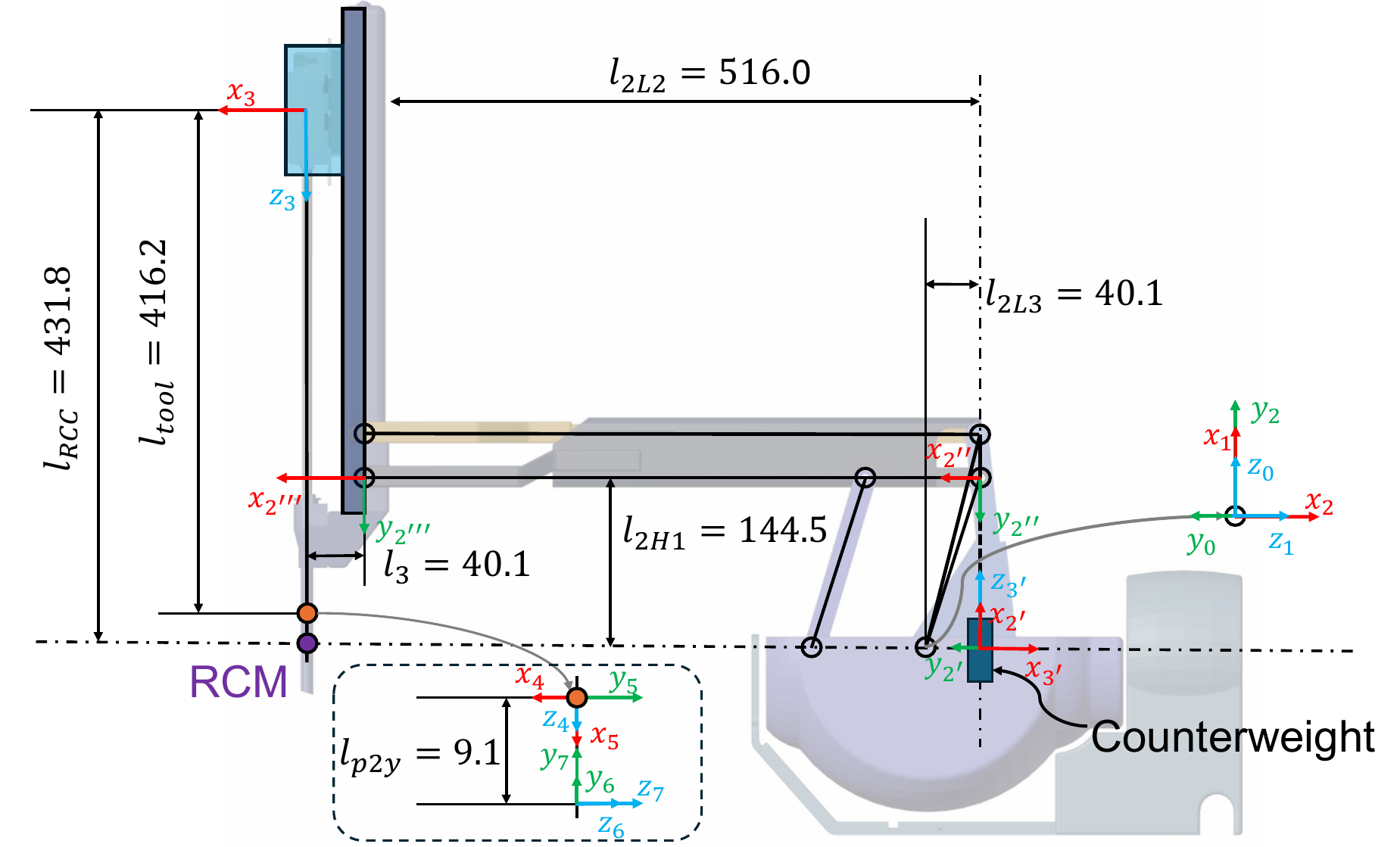}
    \caption{The Planar View of the Frame definition for the PSM. The unit of the link lengths is mm.}
    \label{fig:kinematics}
\end{figure}

The frame definition of the PSM is shown in Fig.~\ref{fig:kinematics}, and the corresponding parameters are shown in Table~\ref{tab:psm_para}. Here we follow the convention from~\cite{wang2019convex} and establish the following relationships among $q, q^d$ and $q^m$:
\begin{subequations}
\begin{equation}
  q_{1-5} = q^d_{1-5},\; q^d_{1-4} = q^m_{1-4}
\end{equation}   
\begin{equation}
  \begin{split}
        \begin{bmatrix} q^d_6 & q^d_7 \end{bmatrix}^T = \begin{bmatrix} \frac{q_6 + q_7}{2} & (-q_6+q_7) \end{bmatrix}^T
    \end{split}
\end{equation}
\begin{equation}
    \footnotesize
    \begin{split}
        \begin{bmatrix} q^d_5 \\ q^d_6 \\ q^d_7 \end{bmatrix} = A_m^d \begin{bmatrix} q^m_5 \\ q^m_6 \\ q^m_7 \end{bmatrix} = \begin{bmatrix} 1.0186 & 0 & 0 \\ -0.8306 & 0.6089 & 0.6089 \\ 0 & -1.2177 & 1.2177 \end{bmatrix} \begin{bmatrix} q^m_5 \\ q^m_6 \\ q^m_7 \end{bmatrix}
    \end{split}
\end{equation}
\end{subequations}




\begin{table}[]
    \begin{center}
    \caption{DH table of the PSM modified from~\cite{wang2019convex}. Two passive joints of the parallelogram mechanism of the second joint removed.}
    \normalsize
    \begin{tabular}{l l l c c l}
        \hline
        i  & Ref & $a_{i-1}$ & $\alpha_{i-1}$ & $d_i$ & $\theta_i$ \\
        \hline
        $1$ & $0$ & 0 & $\frac{\pi}{2}$ & 0 & $q_1 + \frac{\pi}{2}$ \\
        $2$ & $1$ & 0 & $-\frac{\pi}{2}$ & 0 & $q_2 - \frac{\pi}{2}$ \\
        $2'$ & $2$ & $l_{2L3}$ & 0 & 0 & $\frac{\pi}{2}$ \\
        $2''$ & $2'$ & $l_{2H1}$ & 0 & 0 & $\frac{\pi}{2} - q_2$ \\
        $2'''$ & $2''$ & $l_{2L2}$ & 0 & 0 & $q_2$ \\
        $3$ & $2'''$ & $l_{3}$ & $-\frac{\pi}{2}$ & $q_3 + l_{c2}$ & 0\\
        $3'$ & $2$ & $l_{2L3}$ & $-\frac{\pi}{2}$ & $q_3$ & 0 \\
        $4$ & $3$ & 0 & 0 & $l_{tool}$ & $q_4$ \\
        $5$ & $4$ & 0 & $\frac{\pi}{2}$ & 0 & $q_5 + \frac{\pi}{2}$  \\
        $6$ & $5$ & $l_{p2y}$ & $-\frac{\pi}{2}$ & 0 & $q_6 + \frac{\pi}{2}$ \\
        $7$ & $5$ & $l_{p2y}$ & $-\frac{\pi}{2}$ & 0 & $q_7 + \frac{\pi}{2}$ \\
        $M_6$ & - & 0 & 0 & 0 & $q_6^m$ \\
        $M_7$ & - & 0 & 0 & 0 & $q_7^m$ \\
        $F_{67}$ & - & 0 & 0 & 0 & $q_6 - q_7$ \\
        \hline
    \end{tabular}
    
    \footnotesize {Note for the table: (1) Ref stands for the reference link frame of link $i$. (2) $a_{i-1}$, $\alpha_{i-1}$, $d_i$ and $\theta_i$ are the modified DH parameters of link $i$. (3) The parameters of link inertia, motor inertia, joint friction and spring stiffness are omitted since they are the same as ~\cite{wang2019convex}. (4) $M_6$ and $M_7$ correspond to the modeling of motors 6 and 7, respectively. $F_{67}$ corresponds to the modeling of the relative motion between links 6 and 7. (5) $l_{c2}=-l_{RCC} +l_{2H1}$.}
    
    \label{tab:psm_para}
    \end{center}
\end{table}
\subsubsection{Dynamic Modeling of the PSM}

For each link $i$, it has the mass $m_i$, the position of the center of mass (COM) $\mathbf{r}_i$ with respect to link frame $i$ and the inertia matrix $\mathbf{I}_i$ with respect to the COM. Given the above parameters, we can calculate the first moment of inertia $\mathbf{M}_{i} \in \mathbb{R}^3$ and the inertia matrix $\mathbf{L}_i \in \mathbb{R}^{3 \times 3}$ with respect to the link frame $i$:
\begin{equation}
    \begin{split}
        \mathbf{M}_i &= m_i \mathbf{r}_i \\
        \mathbf{L}_i &= \mathbf{I}_i + m_i ([\mathbf{r}_i]_{\times})^T [\mathbf{r}_i]_{\times}
    \end{split}
\end{equation}
where $[\cdot]_{\times}$ is the operator to construct a skew-symmetric matrix from a given vector.

Complete details of the dynamic parameters can be found in~\cite{wang2019convex}. Essentially, we use barycentric parameters~\cite{maes1989linearity} to denote link inertia as $\mathbf{\delta}_{Li}$ and other parameters as $\mathbf{\delta}_{Oi}$. We combine all parameter vectors of $n$ joints and obtain the overall dynamic parameters $\mathbf{\delta}$:
\vspace{-2mm}
\begin{equation}
\vspace{-2mm}
    \begin{split}
        \mathbf{\delta} = \begin{bmatrix} \mathbf{\delta}_{L1}^T \; \mathbf{\delta}_{O1}^T \; \cdots \; \mathbf{\delta}_{Ln}^T \; \mathbf{\delta}_{On}^T \end{bmatrix}^T
    \end{split}
\end{equation}
To address dynamic modeling for the dVRK PSM, we use the Euler-Lagrange method~\cite{nakamura1989dynamics} since it can handle joint constraints. The Lagrangian $L$ is calculated by the difference between the kinetic energy $K$ and potential energy $P$ of the robot, i.e. $L= K - P$. Motor inertia, joint frictions, and spring stiffness are not included in $L$ and are modeled separately. The torques $\tau_{Li}^m$, $\mathbf{\tau}_M^m$, $\mathbf{\tau}_f^c$ and $\mathbf{\tau}_s^c$, which are caused by link inertia, motion inertia, joint frictions, and spring stiffness, can be computed following the formulas in~\cite{wang2019convex}.


The overall torques $\mathbf{\tau}^m$ are given by:
\vspace{-1mm}
\begin{equation}
\vspace{-1.5mm}
    \begin{split}
        \mathbf{\tau}^m &= \tau_{Li}^m  + \mathbf{\tau}_M^m(\Ddot{\mathbf{q}}^m) + \pdv{\mathbf{q}^c}{\mathbf{q}^m} [\mathbf{\tau}_f^c(\Dot{\mathbf{q}}^c) + \mathbf{\tau}_s^c(\mathbf{q}^c)] \\
        &= \mathbf{H}(\mathbf{q}^m, \Dot{\mathbf{q}}^m, \Ddot{\mathbf{q}}^m) \mathbf{\delta}
    \end{split}
\end{equation}
To calculate the base parameters, which are a minimum set of dynamic parameters that can fully describe the dynamic model of the robot, we use QR decomposition with pivoting~\cite{gautier1991numerical}. Then, we can introduce a permutation matrix $\mathbf{P}_b \in \mathbb{R}^{n \times b}$, where $n$ is the number of standard dynamic parameters and $b$ is the number of base parameters. Therefore, we can calculate the base parameters $\mathbf{\delta}_b$ and the corresponding regressor $\mathbf{H}_b$ as:
\vspace{-2mm}
\begin{equation}
\vspace{-2mm}
    \begin{split}
        \mathbf{\delta}_b = \mathbf{P}_b^T \mathbf{\delta} ,\; \mathbf{H}_b = \mathbf{H} \mathbf{P}_b
    \end{split}
\end{equation}

\subsubsection{Excitation Trajectory Optimization}
Periodic excitation trajectories based on Fourier series [24] are used to generate data for dynamic model identification. These trajectories minimize the condition number of the regression matrix $\mathbf{W}_b$ for the base parameters $\mathbf{\delta}_b$ subject to the lower and upper boundary constraints of joint positions \& velocities and the robot Cartesian positions.
\vspace{-1mm}
\begin{equation}
    \begin{split}
        \mathbf{W}_b = \begin{bmatrix} \mathbf{H}_b(\mathbf{q}_1^m, \Dot{\mathbf{q}}_1^m, \Ddot{\mathbf{q}}_1^m),  \cdots,  \mathbf{H}_b(\mathbf{q}_{n_s}^m, \Dot{\mathbf{q}}_{n_s}^m, \Ddot{\mathbf{q}}_{n_s}^m) \end{bmatrix} ^T
    \end{split}
\end{equation}

where $\mathbf{q}_i^m$ is the motor joint coordinate at $i_{th}$ sampling point and $n_s$ is the number of sampling points. The joint coordinate $q_j^m$ of motor $j$ can be calculated by
\begin{equation}
    \small
    \begin{split}
        q_j^m(t) = q_{oj}^m + \sum_{k=1}^{n_H}[\frac{a_{jk}}{k\omega_f}\sin{(k\omega_f t)} - \frac{b_{jk}}{k\omega_f}\cos{(k\omega_f t)}]
    \end{split}
    \label{eq:gen_joint}
\end{equation}
where $\omega_f = 2 \pi f_f$ is angular component of the fundamental frequency $f_f$, $n_H$ is harmonic number of the Fourier series, $a_{jk}$ \& $b_{jk}$ are amplitudes of the $j_{th}$-order sine and cosine function, $q_{oj}^m$ is position offset, and $t$ is time.

\subsubsection{Parameter Identification}
To identify the dynamic parameters, we move the robot along the excitation trajectories generated via the method described in the previous subsection. Data is collected at each sampling time to obtain the regression matrix $W$ and the dependent variable vector $\mathbf{\omega}$.
\begin{equation}
    \vspace{-1mm}
    \begin{split}
        \mathbf{W} &= \begin{bmatrix} \mathbf{H}(\mathbf{q}_1^m, \Dot{\mathbf{q}}_1^m, \Ddot{\mathbf{q}}_1^m), \hdots, \mathbf{H}(\mathbf{q}_{n_s}^m, \Dot{\mathbf{q}}_{n_s}^m, \Ddot{\mathbf{q}}_{n_s}^m) \end{bmatrix}^T \\ \mathbf{\omega} &= \begin{bmatrix} \mathbf{\tau}_1^m, \mathbf{\tau}_2^m, \hdots, \mathbf{\tau}_{n_s}^m \end{bmatrix}^T
    \end{split}
\end{equation}
$\mathbf{\tau}_i^m$ is the motor torque at $i_{th}$ sampling point.

Then, we convert the identification problem into an optimization problem~\cite{wang2019convex}, which minimizes the cost function $Q$ with respect to the decision vector $\mathbf{\delta}$:
\begin{equation}
    \begin{split}
        Q = \norm{\mathbf{W} \mathbf{\delta} - \mathbf{\omega}}^2
    \end{split}
\end{equation}



\subsection{Learning-based Trocar Interaction Torque Correction}

The second step of our method adapts the trocar correction proposed in~\cite{Wu2021RobotFE} to the model-based dynamics identification outputs. We insert the robot instrument through a trocar into an abdominal phantom such that the RCM of the robot is at the incision port. The incision port applies interaction force to the instrument shaft and trocar, which can be regarded as an extra constraint to the robot and effectively changes the system dynamics. The free space torque identification in step one is no longer accurate in this case. However, this extra constraint is hard to model, since in the surgical scenario, the patient body wall is mechanically anisotropic. Furthermore, compliance, strain, and stiffness differ between patients. These factors are subject to the incision location, tissue property, patient age, etc.~\cite{Deeken2017MechanicalPO}. Therefore, to compensate for this additional interaction, as shown in Fig.~\ref{fig:tq_pred}, we train a trocar correction network for each of the six joints based on the residual error between the torque estimated by the free space dynamic model and the measured torque. This network accounts for the interaction between the instrument and the body wall but further work is needed to investigate collisions at other points. The input to each network is a window of concatenated measurements of the previous positions and velocities of all joints, and the current free space estimated torque for each joint. Each network is a multi-layer perceptron consisting of two linear layers with 256 hidden dimensions and a ReLU activation layer in between them. We follow the design in~\cite{Wu2021RobotFE} to minimize the network size to ensure the intra-operative data collection time is clinically feasible. The corrected torque is the sum of the network output and free space network estimation.

\subsection{Cartesian Force/Torque Estimation}

 To get the torque acting upon the external environment ($\hat{\tau}_{ext}$), we subtract the free space (FS) torque estimation with trocar correction ($\hat{\tau}_{FS} - \hat{\tau}_{corr}$) from the measured torque ($\tau$) at every time point. Then, we multiply the inverse transpose of the spatial Jacobian $J^{-T}$ with the external torque to derive the external Cartesian force. Note that the spatial Jacobian here is full-ranked.
\begin{equation} \label{eq:force}
    \hat{F}_{ext} = J^{-T}\hat{\tau}_{ext} = J^{-T} (\tau - (\hat{\tau}_{FS} - \hat{\tau}_{corr}))
\end{equation}

\begin{figure}[htbp]
    \centering
    \vspace{2.5mm}
    \includegraphics[width=0.95\linewidth]{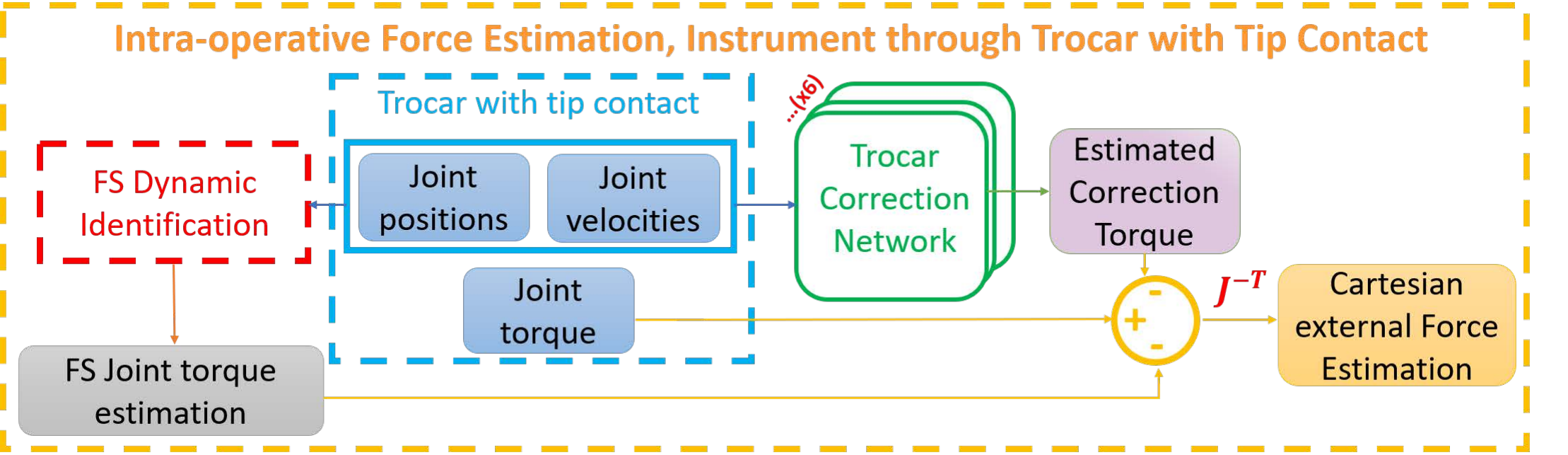}
    \caption{The block diagram of Cartesian force estimation. We use the proposed model to estimate the torque required to achieve the measured joint positions and velocities with no contact. Then, we subtract it from the measured torque and use the Jacobian to convert the difference into force exerted on the environment.}
    \label{fig:force_estimation}
\end{figure}

The force estimation scheme is shown in Fig.~\ref{fig:force_estimation}.

%% file: section/4.Experimental_setup.tex

\subsection{Dynamic Identification and Free Space Torque Estimation}

\subsubsection{Excitation Trajectory Generation}


For each PSM, we generate two independent excitation trajectories for identification and testing purposes. To solve the Eq.~\ref{eq:gen_joint}, we select the harmonic number $n_H$ as 6 and the fundamental frequency $f_f$ of 0.18\,Hz. Furthermore, given the pre-defined joint constraints, we can compute the optimized trajectory using pyOpt\cite{perez2012pyopt}.

\subsubsection{Data Collection and Data Processing}


To collect the data, we run the excitation trajectory with a frequency of 200\,Hz using the dVRK software framework on a dVRK PSM with a Large Needle Driver instrument. We collect the joint positions, velocities, and torques via subscribing to the dVRK Robot Operating System topics~\cite{Quigley2009ROSAO}. Then, we calculate the joint acceleration using the numerical differentiation of the velocities. Lastly, we implement a sixth-order low-pass filter to preprocess the data~\cite{wang2019convex}.

\subsubsection{Solve the model parameters}

To obtain uniformly precise identification results for all joints, we multiply a weight $w_i=\frac{1}{\max{(\tau_i^m)} - \min{(\tau_i^m)}}$ to the cost function $Q_i$ of each motor joint $i$. We optimize the model parameters using the CVXPY package~\cite{diamond2016cvxpy} with the SCS solver~\cite{o2017scs}.

\subsection{Trocar Correction and Cartesian Force Estimation}

We collect about 4 minutes training dataset and split it into 80\%/10\%/10\% as train/validation/test subsets, with a robot instrument inserted through a trocar in an incision port of the abdominal phantom, to learn the correction factor. The correction network is trained for 400 epochs with an initial learning rate of 0.0001, window size of 5, and batch size of 10000. We use Adam and decay learning rate on plateau. Training is done in 15.28 s on an Nvidia A4500 graphic card.

\begin{figure}[]
    \centering
    \vspace{3mm}
    \setlength{\abovecaptionskip}{0.cm}
    \includegraphics[width=0.8\linewidth]{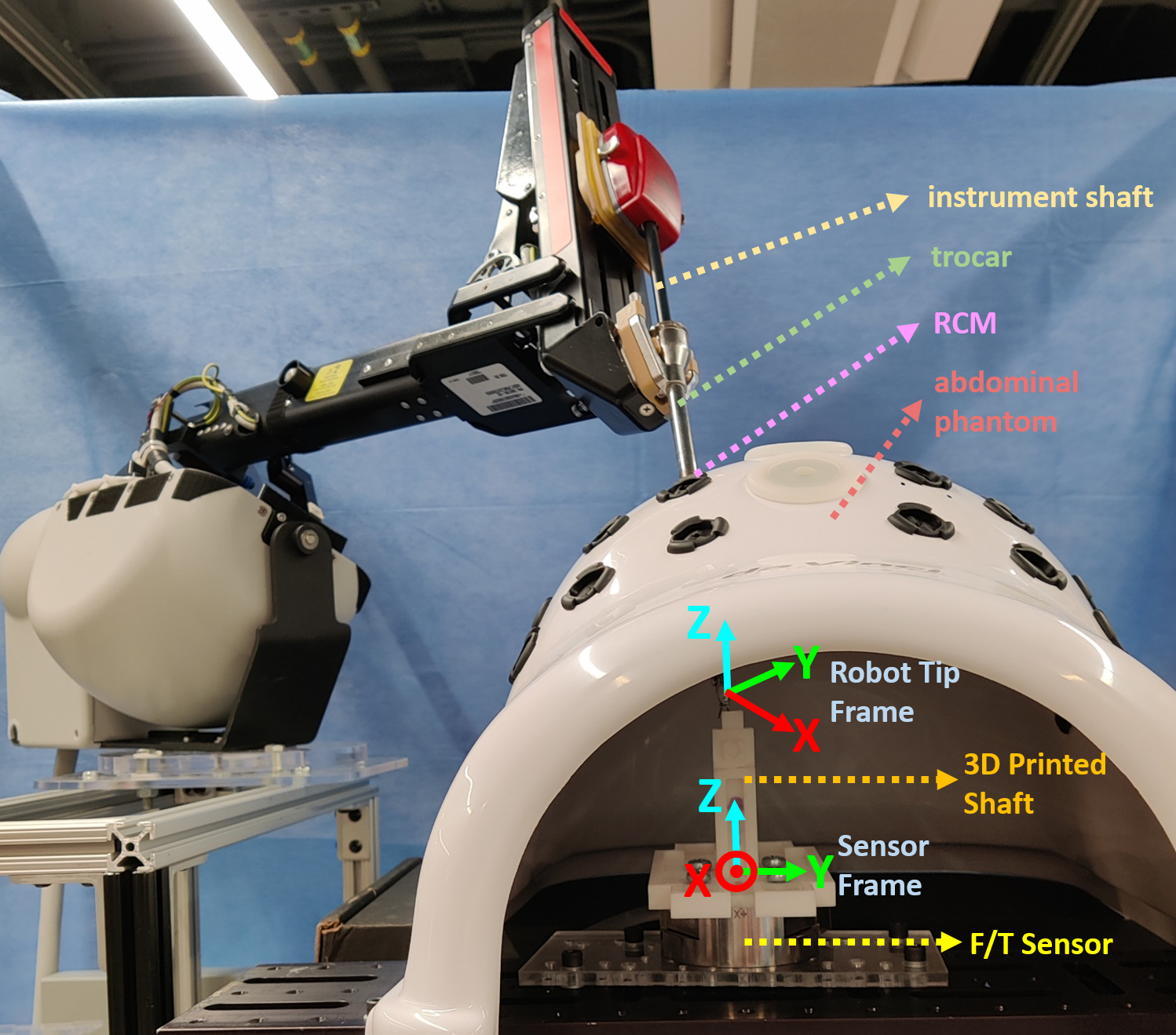}
    \caption{Experimental Setup. The instrument shaft is inserted through a trocar to the abdominal phantom cavity. The trocar is plugged into an incision port on the phantom, which is also the RCM of the robot in this configuration. The instrument tip manipulates the 3D-printed shaft and the force/torque sensor beneath it logs the force that the gripper exerts. Note that there is a 45-degree angle bias between the robot tip frame and the sensor frame, with respect to the sensor frame Z-axis.}      
    \label{fig:exp_setup}
\end{figure}

We then collect an in-trocar, with tip contact test dataset for Cartesian force estimation. We place a Gamma Force/Torque sensor (ATI Industrial Automation, Apex, NC, USA) in the workspace of the robot and mount a 3D-printed shaft on top. We teleoperate the instrument to poke and manipulate the shaft to log external six degrees-of-freedom Cartesian force data, as the ground truth, to compare with our method estimation. The experimental setup is shown in Fig.~\ref{fig:exp_setup}.

We use two metrics, normalized root-mean-square errors (NRMSE) and root-mean-square errors (RMSE) to evaluate the performance of the proposed method. The NRMSE between the actual and estimated joint forces/torques are calculated using the formula given in~\cite{pique2019dynamic}.

%% file: section/5.Experiment.tex

We conduct experiments to compare the force estimation accuracy of our proposed hybrid method with the model-based-only and learning-based-only methods\cite{wang2019convex, Wu2021RobotFE}. We perform free space torque estimation when there is training-testing dataset mismatch, and both torque estimation and Cartesian force estimation when the instrument is in trocar.

\subsection{Learning-based Method Suffers from Train, Test Data Mismatch}

Our proposed hybrid method combines the robustness provided by the physical feasibility from the model-based part, and the adaptability provided by the data-driven compensation from the learning-based part. To validate this, we show that the pure learning-based method can be less reliable in certain conditions. Note that the pure learning-based method proposed in~\cite{Wu2021RobotFE} relies highly on the input training data quality, i.e. whether the robot joint profile in the training dataset can generalize to and cover the workspace in the test dataset. We collect a 15-minute training dataset and partition it into 80\%/20\% train/validation sets. Next, we collect a test dataset in a different workspace.

\begin{figure}[htbp]
    \centering
    \vspace{3mm}
    \setlength{\abovecaptionskip}{0.cm}
    \includegraphics[width=0.98\linewidth]{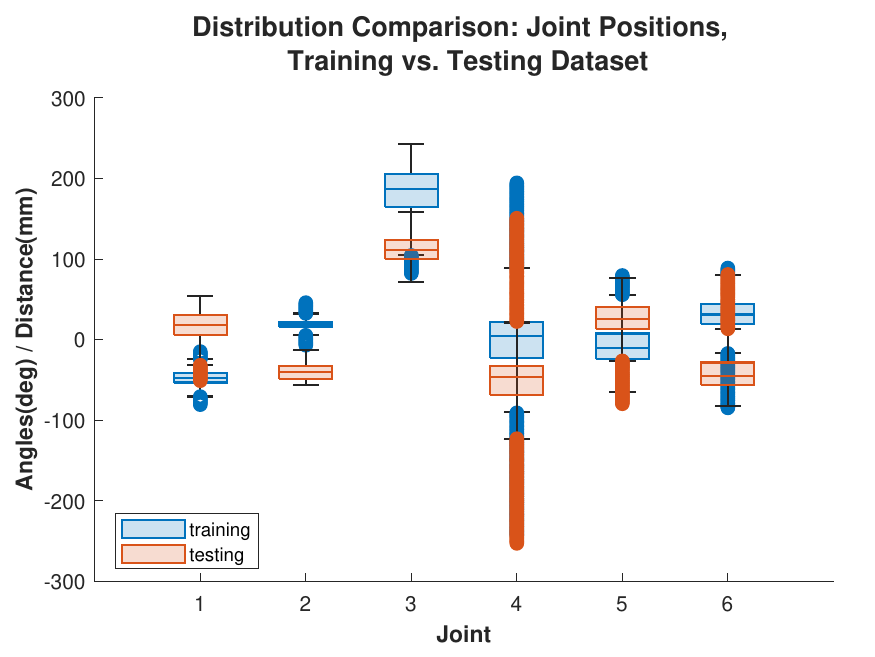}
    \caption{Joint position profile distribution of training and testing dataset collected in different workspace, which causes a mismatch.}  
    \label{fig:bad_dist}
\end{figure}

\begin{figure}[htbp]
    \centering
    \setlength{\abovecaptionskip}{0.cm}
    \includegraphics[width=0.98\linewidth]{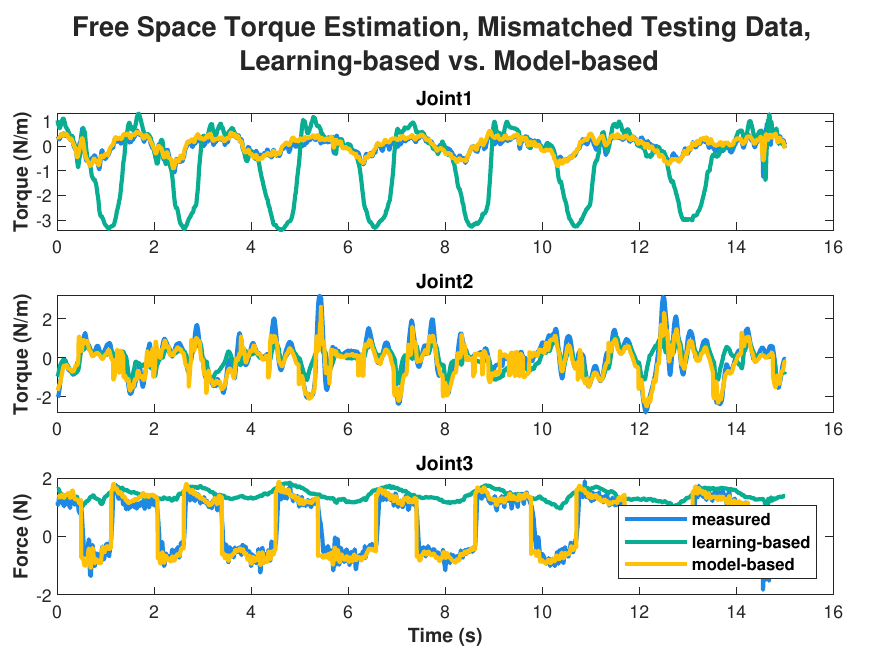}
    \caption{Learning-based vs. model-based free space torque estimation. Since there is a mismatch between training and testing datasets, the learning-based method has high error. Model-based method estimates well, as the dynamic parameters are identified following the proposed scheme.}      
    \label{fig:tc_mf_pm_fe}
\end{figure}

Fig.~\ref{fig:bad_dist} visualizes the different workspace of the joint positions in the training and testing datasets. As our step 2 network is too simple to capture the full workspace, we use the Long Short-Term Memory network proposed in~\cite{Wu2021RobotFE} in this section for a fair comparison with the model-based method. Finally, we estimate torque for the test dataset using the learning-based and optimization-based models, respectively. Results are visualized in Fig.~\ref{fig:tc_mf_pm_fe}, and the RMSE is measured for estimation accuracy evaluation, shown in Table.~\ref{tab:bad_tq_pred}.

The results show that the learning-based method performs poorly when the training and test datasets are not in the same workspace. The training dataset needs to be well-crafted to guarantee test set torque estimation accuracy. 

\begin{table}[htbp]
\centering
\vspace{3mm}
\caption{RMSE of Joint Torque$(N\cdot m)$/Force$(N)$ estimation: Learning-based vs. Model-based with Mismatched Training and Test Data}
\label{tab:bad_tq_pred}
\begin{tabular}{cccc}
\hline
\multicolumn{1}{|c|}{\textbf{Method}} & \multicolumn{1}{c|}{$\tau_1$} & \multicolumn{1}{c|}{$\tau_2$} & \multicolumn{1}{c|}{$f_3$} \\ \hline\hline
\multicolumn{1}{|c|}{Learning-based} & \multicolumn{1}{c|}{0.84} & \multicolumn{1}{c|}{0.49} & \multicolumn{1}{c|}{1.38}  \\ \hline
\multicolumn{1}{|c|}{Model-based} & \multicolumn{1}{c|}{0.22} & \multicolumn{1}{c|}{0.39} & \multicolumn{1}{c|}{0.42}  \\ \hline
\end{tabular}
\end{table}

In contrast, the parameters identified through the model-based method are more generalizable than the learning-based method. Unlike the learning-based method, the model-based method considered the physical feasibility of the robot. Therefore, compared to the uncertainty that the training dataset brings in, the performance of torque estimation by model-based, optimized excitation trajectory is more robust.

\subsection{Phantom Force Estimation: Model-based vs. Hybrid}

\begin{figure}[htbp]
    \centering
    \vspace{3mm}
    \setlength{\abovecaptionskip}{0.cm}
    \includegraphics[width=0.98\linewidth]{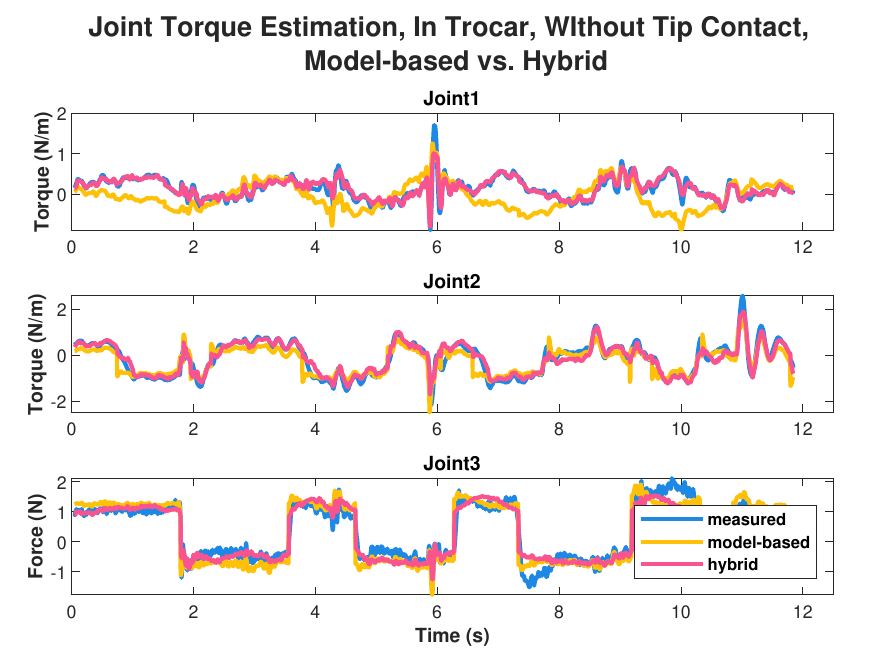}
    \caption{Model-based vs. Hybrid torque estimation with trocar-only contact. Note that the hybrid method contributes the most significant correction on joints 1 and 2. That is because the trocar interaction force mainly exists on the tangent plane of the incision port. Therefore, along the norm of the port, which is the force direction of joint 3, the correction works not as effective as the other two.}      
    \label{fig:to_mb_pm_pd}
\end{figure}

\begin{figure}[htbp]
    \centering
    \setlength{\abovecaptionskip}{0.cm}
    \includegraphics[width=0.98\linewidth]{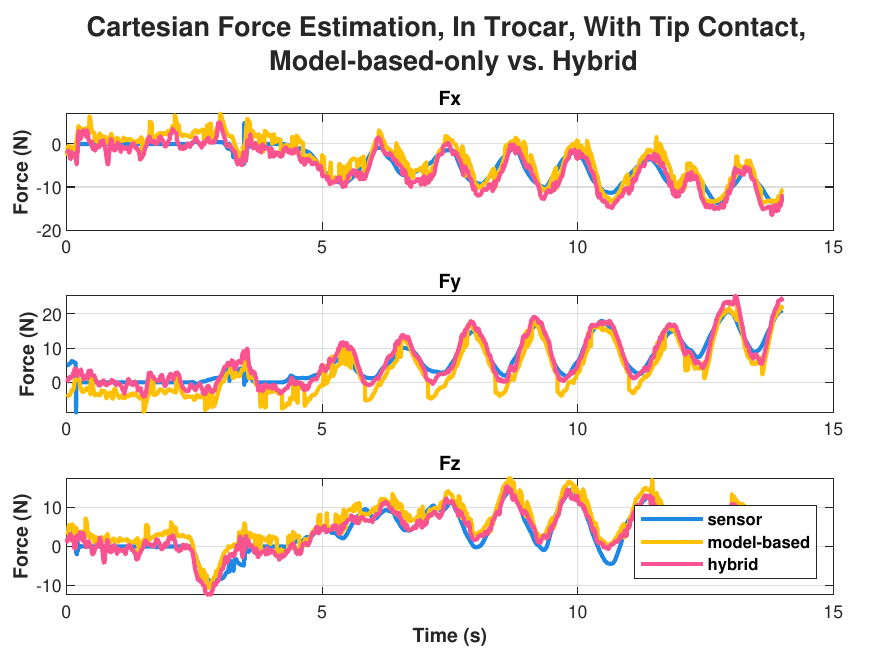}
    \caption{Model-based vs. Hybrid Cartesian Force Estimation, with trocar-abdominal phantom contact and robot tip-F/T sensor contact.}      
    \label{fig:tc_mb_pm_fe}
\end{figure}

We use the experimental setup mentioned in the previous section to achieve trocar correction and Cartesian force estimation and compare the performance with the model-based-only method. We first compare the torque estimation of the in-trocar, no-tip-contact test dataset. Results are shown in Fig.~\ref{fig:to_mb_pm_pd}. Next, we compare the Cartesian force estimation of the in-trocar, with-tip-contact test dataset, as shown in Fig.~\ref{fig:tc_mb_pm_fe}. The NRMSE and RMSE are presented in Table.~\ref{tab:hybrid_tq_pred} and Table.~\ref{tab:hybrid_fc_est}, respectively. The results depict that the model-based method fails to correct extra torque caused by trocar interaction, while the hybrid method does.

\begin{table}[htbp]
\centering
\caption{NRMSE$\dagger$ and RMSE* of In Trocar, No Tip Contact Joint Torque$(N\cdot m)$/Force$(N)$ Estimation: Model-based vs. Hybrid}
\label{tab:hybrid_tq_pred}
\begin{tabular}{ccccccc}
\hline
\multicolumn{1}{|c|}{\textbf{Method}} & \multicolumn{1}{c|}{$\tau_1$} & \multicolumn{1}{c|}{$\tau_2$} & \multicolumn{1}{c|}{$f_3$} & \multicolumn{1}{c|}{$\tau_4$} & \multicolumn{1}{c|}{$\tau_5$} & \multicolumn{1}{c|}{$\tau_6$} \\ \hline\hline
\multicolumn{1}{|c|}{Model-based$\dagger$} & \multicolumn{1}{c|}{0.14} & \multicolumn{1}{c|}{0.07} & \multicolumn{1}{c|}{0.10} & \multicolumn{1}{c|}{0.12} & \multicolumn{1}{c|}{0.35} & \multicolumn{1}{c|}{0.43} \\ \hline
\multicolumn{1}{|c|}{Hybrid$\dagger$} & \multicolumn{1}{c|}{0.067} & \multicolumn{1}{c|}{0.04} & \multicolumn{1}{c|}{0.09} & \multicolumn{1}{c|}{0.09} & \multicolumn{1}{c|}{0.092} & \multicolumn{1}{c|}{0.13} \\ \hline
\multicolumn{1}{|c|}{~\cite{Yilmaz2022TransferOL}$\dagger$} & \multicolumn{1}{c|}{0.062} & \multicolumn{1}{c|}{0.10} & \multicolumn{1}{c|}{0.16} & \multicolumn{1}{c|}{0.06} & \multicolumn{1}{c|}{0.040} & \multicolumn{1}{c|}{0.048} \\ \hline
\multicolumn{1}{|c|}{Model-based*} & \multicolumn{1}{c|}{0.38} & \multicolumn{1}{c|}{0.35} & \multicolumn{1}{c|}{0.45} & \multicolumn{1}{c|}{0.0049} & \multicolumn{1}{c|}{0.07} & \multicolumn{1}{c|}{0.066} \\ \hline
\multicolumn{1}{|c|}{Hybrid*} & \multicolumn{1}{c|}{0.20} & \multicolumn{1}{c|}{0.29} & \multicolumn{1}{c|}{0.44} & \multicolumn{1}{c|}{0.0067} & \multicolumn{1}{c|}{0.03} & \multicolumn{1}{c|}{0.025} \\ \hline
\end{tabular}
\end{table}

\begin{table}[htbp]
\centering
\vspace{3mm}
\caption{NRMSE$\dagger$ and RMSE* of In Trocar, With Tip Contact Cartesian Force Estimation: Model-based vs. Hybrid}
\label{tab:hybrid_fc_est}
\begin{tabular}{cccc}
\hline
\multicolumn{1}{|c|}{\textbf{Method}} & \multicolumn{1}{c|}{$Fx(N)$} & \multicolumn{1}{c|}{$Fy(N)$} & \multicolumn{1}{c|}{$Fz(N)$} \\ \hline\hline
\multicolumn{1}{|c|}{Model-based$\dagger$} & \multicolumn{1}{c|}{0.086} & \multicolumn{1}{c|}{0.098} & \multicolumn{1}{c|}{0.098} \\ \hline
\multicolumn{1}{|c|}{Hybrid$\dagger$} & \multicolumn{1}{c|}{0.062} & \multicolumn{1}{c|}{0.064} & \multicolumn{1}{c|}{0.069} \\ \hline
\multicolumn{1}{|c|}{Model-based*} & \multicolumn{1}{c|}{2.7} & \multicolumn{1}{c|}{3.5} & \multicolumn{1}{c|}{3.3} \\ \hline
\multicolumn{1}{|c|}{Hybrid*} & \multicolumn{1}{c|}{1.9} & \multicolumn{1}{c|}{2.2} & \multicolumn{1}{c|}{2.3} \\ \hline
\multicolumn{1}{|c|}{~\cite{Wu2021RobotFE}*} & \multicolumn{1}{c|}{2.3} & \multicolumn{1}{c|}{1.5} & \multicolumn{1}{c|}{3.3} \\ \hline
\end{tabular}
\end{table}

We observe from Fig.~\ref{fig:to_mb_pm_pd} that our proposed hybrid method estimates better than model-based-only method. The NRMSE of our method is in general within 10\%, which is comparable to the results presented in~\cite{Yilmaz2022TransferOL}. We also observe from Fig.~\ref{fig:tc_mb_pm_fe} that our method effectively estimates Cartesian force with tip contact. The RMSE for all directions are around 2 N, which is comparable to the results presented in~\cite{Wu2021RobotFE}.  As~\cite{Yilmaz2022TransferOL, Wu2021RobotFE} used pure learning-based methods that rely heavily on the quality of training data, these results reflect that the proposed method combines the advantages of both the learning-based and model-based methods. Inflection points are difficult to model for all tested methods. We observe that both the model and NN underestimate large changes and tend to optimize for average result. Lastly, the NN in the correction network may suffer from the same generalization issues that we observe in the full workspace. We posit that it is easier to collect high quality data for the correction network as the workspace is restricted compared to the other pure learning-based methods in~\cite{Yilmaz2022TransferOL, Wu2021RobotFE}. As patient body wall interactions are difficult to model, particularly as data must be collected intraoperatively, the NN is a reasonable trade-off. Additionally, patient body wall interactions may change throughout surgery as insufflation level changes and the NN has the potential to adapt through life-long learning.

%% file: section/6.Conclusion.tex
In this paper, we introduce a hybrid force estimation framework, combining model-based free space torque estimation, learning-based trocar interaction torque correction, and Cartesian force estimation. We show that this method can compensate for the interaction force between the trocar and the patient body, which the pure model-based method cannot address. We also show that this method can be more robust than the pure learning-based method since the model-based trajectory optimization and parameter fitting do not depend on sampling the entire workspace of the robot. In the future, we plan to further investigate the framework universality, such as deploying the methods on other robots. Our hybrid method for sensorless force combines the advantages of model and learning-based methods to provide generalizability and also account for environmental conditions.

\section*{Acknowledgement}
Thanks to Dale Bergman and Alessandro Gozzi for all the hardware support, to Peter Kazanzides and Anton Deguet for dVRK guidance. This work was supported in part by an Intuitive Technology Research Grant. 